\documentclass[runningheads]{llncs}

\usepackage[utf8]{inputenc}
\usepackage{xurl}
\usepackage{graphicx}
\graphicspath{ {images/} }

\begin{document}

\title{Get it right: Improving Comprehensibility with adaptable Speech Expression of a Humanoid Service Robot}

\titlerunning{Improving Comprehensibility with adaptable Speech Expression}
%

\author{Thomas Sievers\orcidID{0000-0002-8675-0122} \and
Ralf Möller\orcidID{0000-0002-1174-3323}}

\institute{Institute of Information Systems, University of Lübeck, 23562 Lübeck, Germany
\email{sievers@ifis.uni-luebeck.de, moeller@ifis.uni-luebeck.de}}

\maketitle

\begin{abstract}
As humanoid service robots are becoming more and more perceptible in public service settings for instance as a guide to welcome visitors or to explain a procedure to follow, it is desirable to improve the comprehensibility of complex issues for human customers and to adapt the level of difficulty of the information provided as well as the language used to individual requirements. This work examines a case study using a humanoid social robot Pepper performing support for customers in a public service environment offering advice and information. An application architecture is proposed that improves the intelligibility of the information received by providing the possibility to translate this information into easy language and/or into another spoken language
\end{abstract}


\section{INTRODUCTION}

As self-service technologies (SSTs), which allow customers to access a service without interaction with service staff, become more common in public services, the use of service robots, which can physically replace a human service worker, offers the opportunity to increase the level and availability of customer service while reducing costs \cite{b1}. Especially in times of an increasing shortage of professionals, the use of robots appears to be a practical way to relieve skilled workers. Moreover, in an unsettled world situation the requirements for dealing with people with a different cultural background and lack of language skills are increasing, especially for the authorities in public services. Therefore, it would be desirable that robots in this field are able to interact effectively with human clients across different cultural settings. These robots need social and cross-cultural skills to provide information and assistance in an adequate way. They should be able to respond to the client's linguistic and intellectual abilities.

Most of the people surveyed in a study by TU Darmstadt on the robotization of office and service professions \cite{b2} would accept robots as counterpart in a variety of service environments. More than 80 per cent of respondents, however, prefer human contact for sensitive, personal services such as complex financial advice, psychological or medical care. It is therefore important that a robot addresses people and engages with them in an appropriate manner, not least because many people are still surprised that they can communicate with a robot in a natural way using natural language \cite{b3}. Robots will be most successful in being accepted if they meet expectations through aesthetic characteristics that engage users within a clearly defined context \cite{b4}. Everything from the robots appearance to the visual design and the sound of voice hold semantic value. Previous work has also provided insights into the effects of language and cultural context on the credibility of robot speech \cite{b5}. The way robots express themselves to build credibility and convey information in a meaningful and compelling way is a key function in creating acceptance and usability.

A decisive factor for the success of the use of service robots in interaction with customers is the customers' trust or lack of trust respectively \cite{b6}. Trust is a multidimensional concept, reflecting the perceived competence, integrity and benevolence of another entity \cite{b7} and therefore a strong determinant of intention for customers to use a service robot. If they do not trust it they will not enjoy using it and therefore the potential of service robots is not exploited. In research a lack of trust in service robots is often cited as the main reason for customers not to have the intention to use the service robot \cite{b8} \cite{b9}. Establishing trust and trying to create empathy toward the robot while the customers believe in the robot having empathy towards them is considered indispensable for a successful service and caring process \cite{b4}.

In their research van Pinxteren et al. \cite{b6} focus on anthropomorphism as a central concept in human-robot interaction (HRI). According to theory, human-like features of the robot facilitate anthropomorphism \cite{b10}. More than 80 per cent of respondents surveyed by TU Darmstadt \cite{b2} believe that robots can show feelings. One recalls the Arthur C. Clarke quote that "any sufficiently advanced technology is indistinguishable from magic“, and according to Diana \cite{b4} this phenomenon makes it all the more important for a robot creator to build a strategy that acknowledges limitations helping people to understand what to expect. An enlightenment of the average user assuming magical, sophisticated knowledge and perception of the robot will help to avoid miscommunication. It must be clear and understandable to the customer what can be expected from the service robot.

People tend to blame robots, when they experience them as social actors, for mistakes and annoyances to the degree they attribute to their autonomy \cite{b11}. Thus, especially when something fails in completing the desired task for example due to misunderstandings, the human perception of the service robot can be a critical component for the acceptance of robotic services in general. Greater autonomy of the robot can lead to greater blame if something goes wrong.

This paper discusses an approach to making communication between service robots and human clients easier to understand in order to reduce possible fears, barriers and misunderstandings and to enable successful and goal-oriented communication. We propose features for a humanoid social robot that can improve its intelligibility and thus the acceptance of such a service robot in environments that are sensitive and anxiety-provoking for humans.

\section{PRELIMINARIES}

Especially in the public service, there is a need to facilitate access to important information. In Germany alone, more than 10 million people \cite{b12} (some speak of more than 17 million people \cite{b13}) require text in easily understandable language to access crucial information and successfully manage their daily lives. There are many different names and designations for easy language: Easy language, simple language, easily understandable language, barrier-free information and communication \cite{b14}. Easy language is defined as a language style with simpler sentence structure and additional explanations. The target group that can benefit from texts in easy language include people with learning difficulties, little knowledge of administrative language, reading difficulties, older people or people suffering from dementia, as well as people learning the official language as a second language. This results in the need to provide as much information as possible in easy language.

In order to avoid misunderstandings, it would also be a great advantage if the relevant content were offered in a language that the client understands well. This would usually be the client's mother tongue. Clients who turn to public services have a specific problem they want to solve and often do not know how to achieve the desired result. They are therefore in a state of uncertainty. In this state of mind, it is very easy to get upset when things are not going in the right direction, and then misunderstandings, angry reactions, and agitation often occur. To avoid this, it is better to communicate with clients in their own language because they can express themselves more easily, important bits of information won't get lost in translation and the client’s anxiety of speaking another language is removed. If clients or customers are upset, they might not want the extra mental effort of communicating in a second language. As multicultural expert Michael Soon Lee states "Even if you only have a basic command of the customer’s native language, it may take longer to get your point across, but the customer is much more likely to be receptive and understanding" \cite{b15}.

The eGovernment MONITOR 2022 \cite{b16} shows that only 54 percent of the population in Germany uses eGovernment services at all. More than half of all citizens still use services for which they have a need in analog form (57 percent). Lack of knowledge of online availability is the central reason for the digital usage gap for many services. Under these circumstances, it cannot be assumed that advance information and preparation for a visit to a public authority with regard to necessary documents, requirements, etc. is always optimal. This is probably even more true for people with a migration background.

Misunderstandings, ignorance, and lack of understanding on the part of cliens can also create stress for public service staff to deal with. Therefore, conflicts cannot be ruled out and a friendly and approachable interaction with each other is sometimes difficult. Despite its human-like appearance, the Pepper robot remains a little abstract and does not resemble a real human being in all facets, for example in a sometimes all too human unkind reaction to incomprehension. The service robot always remains calm, friendly and courteous, whatever happens. These skills make him a valuable service provider for clients when supporting service staff.

This leads to the conclusion that humanoid service robots equipped with the flexibility to present information in easy language and/or in another language if needed can be of great benefit to people seeking advice, information or help, especially if these people come from a different country or have difficulties in understanding for other reasons. The improvements we propose in the intelligibility and usability of public services are difficult to realise by human agents, but we will show that it is relatively easy to equip a robot with these features.

\section{RELATED WORK}

We have not found many implementations for the use of easy language or translation by a humanoid robot to improve the comprehensibility of information for human counterparts. In particular, the use of easy language does not seem to have been the subject of research related to humanoid robots so far. A robot-bound speech-to-speech translation system is proposed in the domain of medical care as a one-way translation designed to help English speaking patients describe their symptoms to Korean doctors or nurses \cite{b17}. The use of a social robot to teach vocabulary to Turkish-Dutch children and the extent to which a bilingual robot or a monolingual robot promotes word learning is investigated by Leeuwestein et al. \cite{b18}. Santano \cite{b19} shows how to provide Pepper with translating capabilities using Google's ML Kit Translation API \cite{b20}. This works in offline mode as on-device translation and is intended for casual and simple translations only as it does not offer the same quality as Google's Cloud Translation API.

Meaningful and communicative robot actions expressed through body language interpretable by humans offer a further channel of communication with measurable effects on the interaction \cite{b21}. This implicit communication through movement has the potential to increase efficiency in collaborative tasks between robots and humans and is therefore very promising in terms of social interaction.

A reference framework adding intelligibility to the behavior of a robotic system for improvement of predictability, trust, safety, usability, and acceptance of autonomous robotic systems is proposed by van Deurzen et al. \cite{b22}. It  comprises an interactive, online, and visual dashboard to help identify where and when adding intelligibility to the interface design is required so that developers and designers can customise the interactions to improve the experience for people working with the robot.

These approaches help to improve communication between humans and robots, or between humans with the help of robots, without simplifying the complexity of the language or the representation of the information itself. However, in such simplification lies a key to better understanding.

\section{HUMANOID ROBOT PEPPER}

\begin{figure}
  \includegraphics[width=\textwidth]{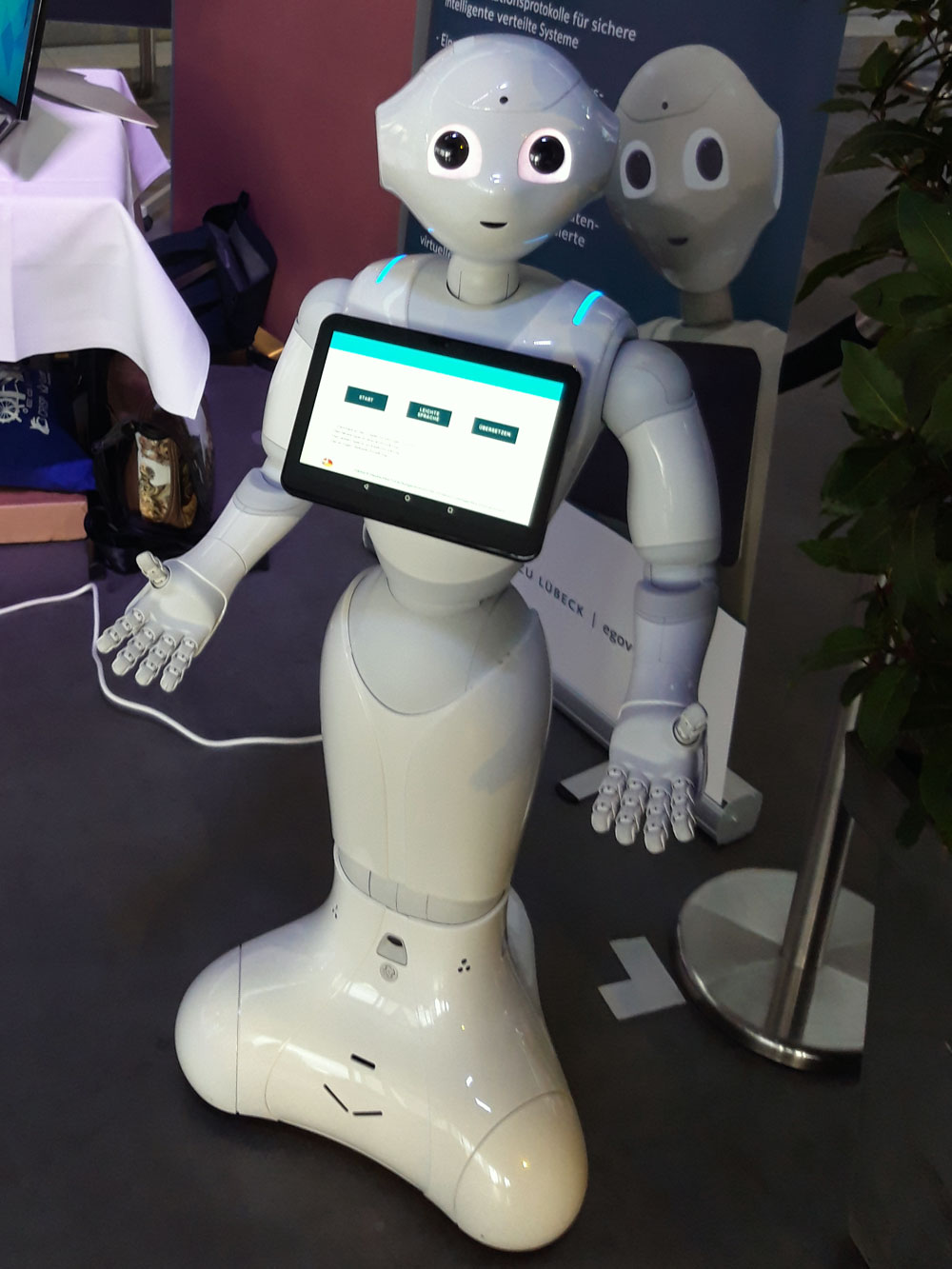}
  \caption{Humanoid Robot Pepper}
  \label{fig_pepper}
\end{figure}

The social humanoid robot Pepper \cite{b23} as seen in Figure~\ref{fig_pepper} was developed by Aldebaran and first released in 2015. The robot is 120 centimeters tall and optimized for human interaction. It is able to engage with people through conversation, gestures and its touch screen. Pepper is equipped with internal sensors, four directional microphones in his head and speakers for voice output. The robot is able to process images from its 3D camera and two HD cameras by shape recognition software for identification of faces and objects so that it can focus on, identify, and recognize people. Speech recognition and dialogue is available in 15 languages. Beyond, Pepper can perceive basic human emotions.

The robot features an open and fully programmable platform so that developers can program their own applications to run on Pepper using software development kits (SDKs) for programming languages like C++, Python or Java respectively Kotlin. This approach allows the development of robot applications for a wide variety of scenarios in a development environment familiar to most developers.

On Pepper's chest a tablet is located presenting a screen resolution of 1280 by 800 pixels. This tablet, which runs Google's Android system, can be equipped with applications to create interaction between the robot and a human user via a straightforward application programming interface (API). There is a Pepper SDK \cite{b24}, which is an Android Studio plug-in and provides a set of graphical tools and a Java library when using Android Studio as an integrated development environment (IDE). The Pepper SDK contains an emulator for the tablet and a robot viewer that shows, among other things, a dialogue view with which the robot's speaking actions can be tested without needing the real robot.

Since research has generally shown that trust is the basis for successful communication tasks and trust in robots is increased by anthropomorphism, a humanoid social robot like Pepper is a good choice for service delivery in interaction with customers. A human face, the possibility of human-like expressions and body language, the use of voice and a name of its own are seen as beneficial for the trust of customers in the robot \cite{b25}.

\section{CASE STUDY}

Our case study relates to the scenario of a customer centre of a public service authority, where many people regularly pursue different concerns such as applying for identity cards, residence permits or work permits. Official procedures usually require knowledge about necessary requirements, documents needed, etc. Many procedures also require explanation for people who rarely have to deal with them, especially if they have a limited understanding of the language and/or belong to a different cultural group with different experiences and background. This can easily lead to uncertainties and tense situations that make it difficult to process the respective case together with the case worker in a stress-free and relaxed manner. Furthermore, it is unfavourable if, for example, the lack of prerequisites or necessary documents is noticed only during the conversation with a human case worker. It would be better if an applicant was informed about everything necessary beforehand and in a way that is understandable and comprehensible to him or her. This would save frustration and unnecessary waiting time on the part of the customer, as well as needless time spent by the case worker on cases that cannot be processed directly due to a lack of sufficient prerequisites.

Such information and education about necessary prerequisites can be provided by a robot with social interaction capabilities. This robot, positioned in the waiting or reception area of an authority with personal customer contacts, can approach individual customers and ask about their concerns. Equipped with application-specific expertise, the robot can provide helpful information and clarification on the respective issue. Furthermore, a humanoid social robot can also provide additional general guidance and orientation in such a scenario.

This paper discusses a possible approach to automatically reduce the complexity of regulatory information or adapting it to the needs of individual customers by translating such content into easy language and/or a foreign language. We think that with such measures, an improvement of adequate information of the customers in terms of comprehensibility and preparation possibilities can be achieved, as well as an increase in the effectiveness of the processes in the conversation with the clerk. This would result in time savings, better preparation, reduction of anxieties and perhaps a more relaxed social intercourse.

\section{APPLICATION ARCHITECTURE}

\begin{figure}
  \includegraphics[width=\textwidth]{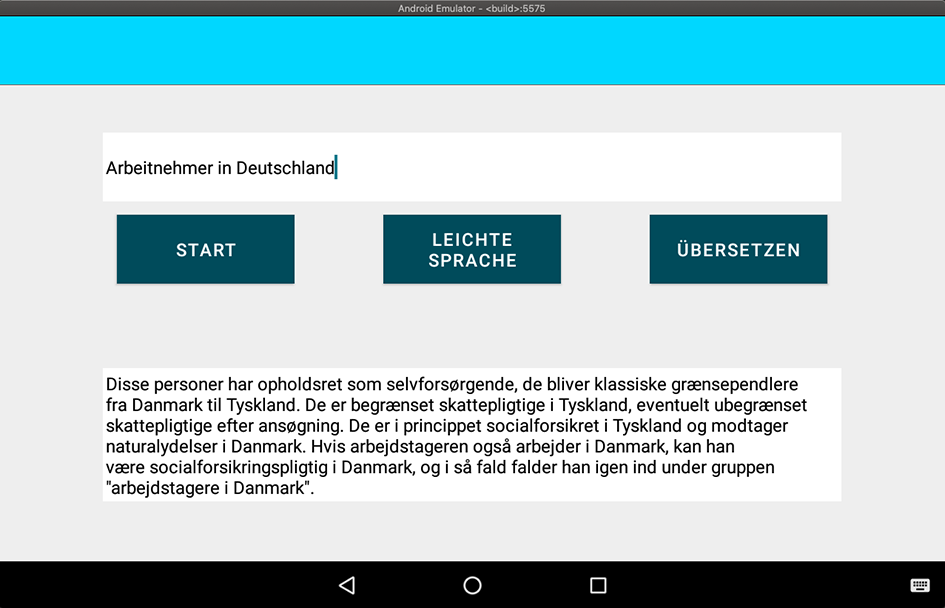}
  \caption{Tablet as user interface displaying obtained information translated to Danish}
  \label{fig_tabletTranslation}
\end{figure}

We developed our exemplary application to achieve this goal using Android Studio and the programming language Kotlin. We use the Pepper SDK for Android \cite{b24} which allows the robot to be controlled from our Android application on its tablet with existing functionalities of Pepper's operating system for focusing upon a person, listening, talking and chatting as well as movements of head and arm to underline what is said with appropriate movements. Every output the robot performs by voice is also displayed on its tablet for the purpose of redundancy serving those with hearing impairment. The tablet serves as a user interface showing the obtained information and offering the possibility to make a selection for easy language or translation as can be seen in Figure~\ref{fig_tabletTranslation}. This selection can also be made by voice using a phrase containing predefined key words like "easy language" or "translation". A blue speech bar on the tablet indicates listening and an animation is shown when Pepper is receiving and processing information.

As a knowledge base in our case study for the case-specific complex expert knowledge that the robot should be able to provide information about, we use data from a MySQL database hosted on a web server with a self-programmed application programming interfaces (API) that returns text answers in JSON (JavaScript Object Notation) format. This API transmits a term or phrase that the robot has heard and recognised to the server via Wi-Fi connection. A PHP script running on the web server then returns a corresponding response from the database. This basic example knowledge base could be something totally different in another case of application. It represents the initial information that needs to be simplified.

For translation into easy language and into a foreign language, we use services that are hosted on various external servers and are not proprietary. These services are also connected to the robot application via Wi-Fi and a corresponding API connection. We connect to the AI-powered translation tool for easy language of SUMM AI \cite{b26} and use the API of DeepL \cite{b27} for translation services. Text given to this APIs is automatically returned in a JSON format that is handed over to the voice and tablet output of the robot. This method and the use of the APIs is described in more detail by Sievers et al. \cite{b28}. SUMM AI currently supports easy language for German. DeepL provides translation to 27 languages. In our proof-of-concept application, we only use one foreign language, Danish. The Pepper robot has to be equipped with the appropriate language package for being able to pronounce used languages correctly.

\subsection{Application Structure}

\begin{figure}
  \includegraphics[width=\textwidth]{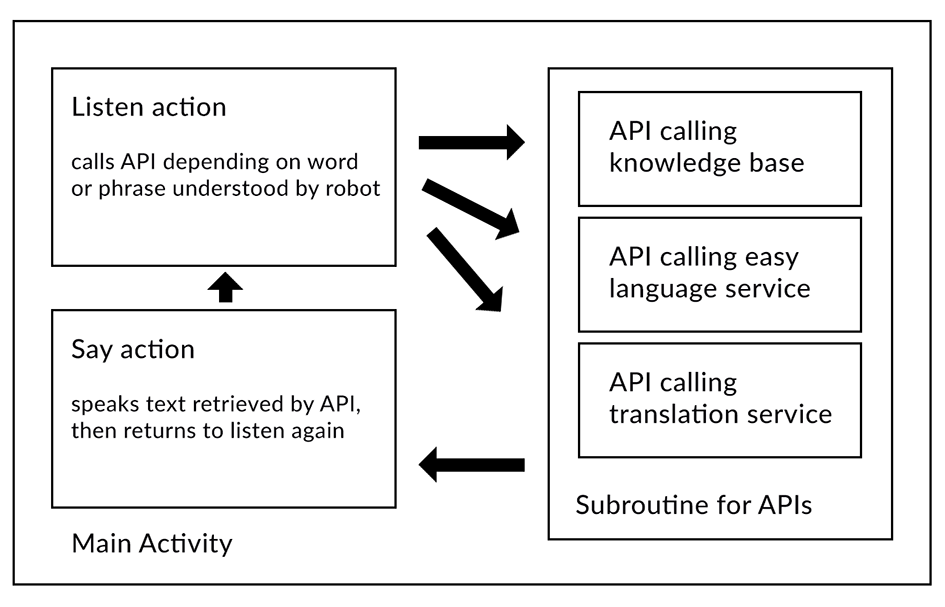}
  \caption{Main Activity of our application}
  \label{fig_mainActivity}
\end{figure}

The Android application with its \textit{Main Activity} acts as a scaffold and control unit for the basic proceedings. It contains the program routines necessary for the usage of native robot resources like listening and talking and subroutines for calling the APIs as illustrated in Figure~\ref{fig_mainActivity}. A subroutine for asking a customer after 10 seconds idling if the service robot could help with another topic is also implemented.

\subsection{Listen \& Say or Chat}

Our first approach was to use the Pepper-specific \textit{Listen} and \textit{Say} functionalities \cite{b29} \cite{b30} to realise a dialogue between robot and customer. The Pepper robot is able to react to words or phrases provided in the program code of the application. Real speech recognition in a general sense is not possible with Pepper's native abilities but the robot is able to spot a word or phrase known to him while listening. Our application contains the necessary vocabulary for the words and phrases that the robot should understand, divided into different phrases for different topics. These topics refer to the knowledge base stored in our database. For every topic we provide a set of words or phrases serving as keywords for the retrieval of the related information. Positively understood words or phrases are forwarded to a subroutine that processes the first API call to the knowledge base.

\begin{figure}
  \centering
  \includegraphics[width=10cm]{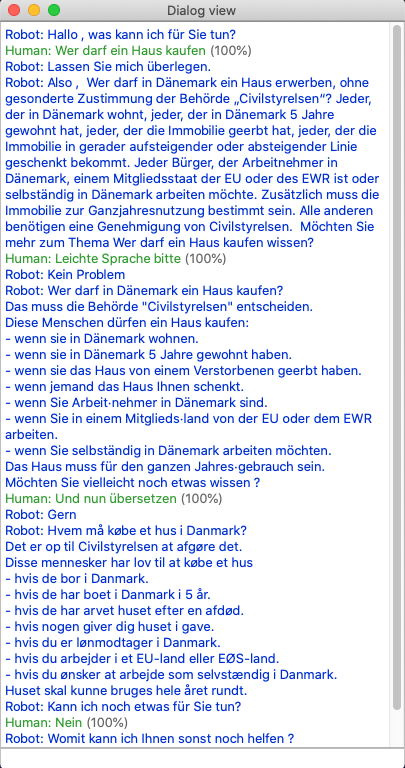}
  \caption{Robot emulation viewer showing dialogue between robot and human with knowledge base answer, translation to easy language, and translation to Danish}
  \label{fig_emulatorChat}
\end{figure}

In a second approach, we used Pepper's \textit{Chat} feature \cite{b31} to conduct the dialogue. This was found to bring a significant improvement in terms of spoken language recognition. Using \textit{Listen} action the word or phrase of the topic is mostly only understood if it is said apart from the context of other words or a sentence. With the \textit{Chat} function, individual words and short sentences of a topic are usually understood even if they are spoken as part of a longer sentence. Also a more natural flow of dialogue is possible using the \textit{Chat} feature due to the flexible possibilities of using variables or randomly selected sentence components in the robot's responses as can be seen in Figure~\ref{fig_emulatorChat}.

By using pauses in speech, emphasis and modulation of the voice, intelligibility can be further improved. The Pepper SDK provides parameters for these purposes that can be used in the dialogue topics. All in all, this opens up the possibility of creating a customer-oriented dialogue on the part of the robot that is characterised by politeness and friendliness.

\subsection{Animation}

\begin{figure}
  \includegraphics[width=\textwidth]{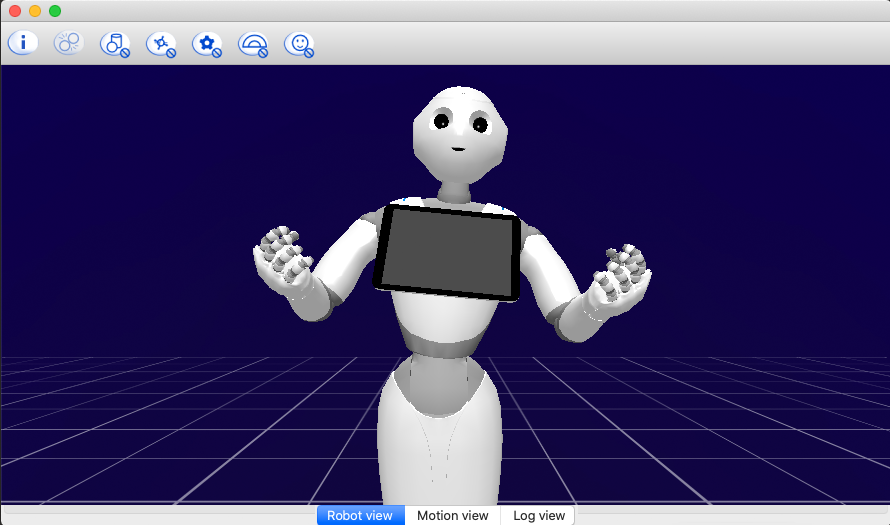}
  \caption{Robot emulation viewer showing the gesture of an animation}
  \label{fig_emulatorAnimation}
\end{figure}

The use of robot animation depending on a specific context is also possible with Pepper. Predefined animations can be easily integrated depending on different events of the dialogue to enhance anthropomorphism and intelligibility through the indirect effect of body language. Creating your own animations would also be possible. In our case study, we have limited the use of animations to existing ones in the Pepper SDK for demonstration purposes. We have defined groups of appropriate animations, from which a randomly selected one is executed at certain points of the interaction, e.g., at the greeting, in response to a question from the human, when the robot asks a question, and so on. Figure~\ref{fig_emulatorAnimation} shows an example of a gesture. These animations support the interaction with the customers as they underline statements of the robot. However, due to the abstraction of the robot's form, the degree of reality of this non-verbal behaviour never reaches a level that would have a negative impact on the acceptance of the robot's behaviour with regard to the \textit{uncanny valley} effect \cite{b32}.

\section{CONCLUSIONS}

With this paper we propose a method to improve the comprehensibility of complex official information for human customers by adapting the level of content-related difficulties. We suggest that such content, when a customer asks for it, be automatically translated into easy language or a foreign language by a humanoid social robot. Our case study involves a customer-facing government agency where a Pepper robot does just that.

Workshops we did together with public service staff show that such features are seen as useful and that one can well imagine the use of such a robot in suitable public authorities. The use of the aforementioned functional APIs via an internet connection proves to be possible without any problems and without unreasonably long response times. An integration of external services via API into applications using the Pepper SDK in Android Studio is quite possible for experienced Android developers. Additional knowledge about the inner workings of the connected services is not required, which is particularly advantageous when integrating complex AI technologies and simplifies application development for such technologies. This approach is not limited to a specific robot model, but can be realised as an API connection for various models.

In future work, we hope to further improve the robot's capabilities in dialogue with human customers and make it less prone to errors. Using language models such as Generative Pre-trained Transformer 3 (GPT-3) or more recent versions to generate human-like text may be an option. Data protection concerns also require closer consideration. 

Through further workshops with authorities and practical tests of our hypothesis in an environment corresponding to our case study, we will seek to gain a better understanding of how best to design our proposed application, overcome any limitations and improve human-robot interaction to achieve the goal of assisting individuals who have difficulty understanding complex content required in public service.

%
%
%
\bibliographystyle{splncs04}

\begin{thebibliography}{8}

\bibitem{b1} G. Allmendinger and R. Lombreglia, "Four strategies for the age of smart services," Harvard Business Review, vol. 83, pp. 131--143, 2005.
\bibitem{b2} R. Stock-Homburg. (2016) Denn sie wissen nicht, was sie tun - Studie der TU Darmstadt zur Robotisierung von Büro- und Dienstleistungsberufen. Transnational Study series "Robots@work4.0" [Online].  Available: \url{https://www.tu-darmstadt.de/universitaet/aktuelles_meldungen/archiv_2/2016/2016quartal4/einzelansicht_162880.de.jsp}
\bibitem{b3} A. Gardecki, M. Podpora, R. Beniak and B. Klin, "The Pepper humanoid robot in front desk application,” Conference Paper: Progress in Applied Electrical Engineering (PAEE), doi: 10.1109/PAEE.2018.8441069, 2018.
\bibitem{b4} C. Diana, "My Robot gets Me: How social design can make new products more human," Havard Business Review Press, Boston, MA, ISBN 9781633694422, pp. 218-228, 2021.
\bibitem{b5} S. Andrist, M. Ziadee, H. Boukaram, B. Mutlu, and M. Sakr, "Effects of Culture on the Credibility of Robot Speech: A Comparison between English and Arabic," Proceedings of the Tenth Annual ACM/IEEE International Conference on Human-Robot Interaction (HRI '15), Association for Computing Machinery, New York, NY, USA, pp. 157--164, \doi{10.1145/2696454.2696464}, 2015.
\bibitem{b6} M.M.E. van Pinxteren, R.W.H. Wetzels, J. Rüger, M. Pluymaekers and M. Wetzels, "Trust in humanoid robots: implications for services marketing“, Journal of Services Marketing 33/4, pp. 507--518, 2019.
\bibitem{b7} R.C. Mayer, J.H. Davis and F.D. Schoorman, "An integrative model of organizational trust“, Academy of Management Review, Vol. 20 No. 3, pp. 709--734, \doi{10.2307/258792}, 1995.
\bibitem{b8} J. Everett, D. Pizarro and M. Crockett. (2017) Why are we reluctant to trust robots? [Online].  Available: \url{https://www.theguardian.com/science/head-quarters/2017/apr/24/why-are-we-reluctantto-trust-robots}
\bibitem{b9} B. Morgan. (2017) 10 Things robots can’t do better than humans [Online]. Available: \url{https://www.forbes.com/sites/blakemorgan/2017/08/16/10-things-robots-cant-do-better-than-humans}
\bibitem{b10} N. Epley, A. Waytz and J.T. Cacioppo, "On seeing human: a three-factor theory of anthropomorphism“, Psychological Review, Vol. 114 No. 4, pp. 864--886, \doi{10.1037/0033-295X.114.4.864}, 2007.
\bibitem{b11} C. Furlough, T. Stokes and DJ. Gillan, "Attributing Blame to Robots: I. The Influence of Robot Autonomy," Hum Factors, Vol. 63 No. 4, pp. 592--602, \doi{10.1177/0018720819880641}, 2021.
\bibitem{b12} Press and Information Office of the Federal Government. (2021) Any complex content can be translated into easy language [Online]. Available: https://www.bundesregierung.de/bregde/aktuelles/interview-anne-leichtfuss-1918176
\bibitem{b13} Quatropus GmbH \& Co. KG. (2022) On the status quo of easy language in Germany [Online]. Available: https://www.quatrolingo.com/status-quo-leichte-sprachedeutschland-2022/
\bibitem{b14} capito. Easy-to-understand language and barrier-free information [Online]. Available: \url{https://www.capito.eu/en/what-is-easy-to-understand-language/}
\bibitem{b15} Michael Soon Lee. Driving Sales and Satisfaction with Multilingual Service [Online]. Available: \url{http://resources.rosettastone.com/CDN/us/pdfs/Biz-Public-Sec/Driving-Sales-and-Satisfaction-with-Multilingual-Service.pdf}
\bibitem{b16} Initiative D21 e. V. (2022) eGovernment MONITOR [Online]. Available: \url{https://initiatived21.de/egovernment-monitor/}
\bibitem{b17} S. Shin, E.T. Matson, Jinok Park, Bowon Yang, Juhee Lee and Jin-Woo Jung, "Speech-to-speech translation humanoid robot in doctor's office," 2015 6th International Conference on Automation, Robotics and Applications (ICARA), pp. 484--489, \doi{10.1109/ICARA.2015.7081196}, 2015.
\bibitem{b18} H. Leeuwestein, M. Barking, H. Sodacı et al., "Teaching Turkish-Dutch kindergartners Dutch vocabulary with a social robot: Does the robot's use of Turkish translations benefit children's Dutch vocabulary learning?" Journal of Computer Assisted Learning, 37, pp. 603--620, \doi{10.1111/jcal.12510}, 2021
\bibitem{b19} S. Santano, innovex GmbH (2023) How to Use Google’s ML Kit to Enhance Pepper With AI (Part 5) [Online]. Available: https://www.inovex.de/de/blog/how-to-use-googles-ml-kit-to-enhance-pepper-with-ai-part-5/
\bibitem{b20} Google. (2023) Translation [Online]. Available: https://developers.google.com/ml-kit/language/translation
\bibitem{b21} L. Lastrico, N. F. Duarte, A. Carfì, F. Rea, F. Mastrogiovanni, A. Sciutti and J. Santos-Victor, "If You Are Careful, So Am I! How Robot Communicative Motions Can Influence Human Approach in a Joint Task," \doi{10.48550/ARXIV.2210.13290}, 2022.
\bibitem{b22} B. van Deurzen, H. Bruyninckx and K. Luyten, "Choreobot: A Reference Framework and Online Visual Dashboard for Supporting the Design of Intelligible Robotic Systems," Proc. ACM Hum.-Comput. Interact. 6, EICS, Article 151, 24 pages. \doi{10.1145/3532201}, 2022
\bibitem{b23} Aldebaran, United Robotics Group and Softbank Robotics. (2022) Pepper [Online]. Available: \url{https://www.aldebaran.com/en/pepper}
\bibitem{b24} Aldebaran, United Robotics Group and Softbank Robotics. (2022) Pepper SDK for Android [Online]. Available: \url{https://qisdk.softbankrobotics.com/sdk/doc/pepper-sdk/index.html}
\bibitem{b25} J. Fink, "Anthropomorphism and Human Likeness in the Design of Robots and Human-Robot Interaction," in International Conference on Social Robotics (ICSR), Chengdu, China, Springer, Heidelberg, Berlin, pp. 199--208, \doi{10.1007/978-3-642-34103-8\_20}, 2012
\bibitem{b26} SUMM AI GmbH. (2022) Summ - easy language [Online]. Available: https://summ-ai.com/en/
\bibitem{b27} DeepL SE. (2022) Translate with the deepl api [Online]. Available: https://www.deepl.com/pro-api
\bibitem{b28} T. Sievers, M. Bender and R. Möller, "Connecting AI Technologies as Online Services to a Humanoid Service Robot," to be published in: 15th International Conference on Computer and Automation Engineering, (ICCAE 2023), March 3-5 2023.
\bibitem{b29} QiSDK. (2022) Listen [Online]. Available: \url{https://qisdk.softbankrobotics.com/sdk/doc/pepper-sdk/ch4\_api/conversation/reference/listen.html}
\bibitem{b30} QiSDK. (2022) Say [Online]. Available: \url{https://qisdk.softbankrobotics.com/sdk/doc/pepper-sdk/ch4\_api/conversation/reference/say.html}
\bibitem{b31} QiSDK. (2022) Chat [Online]. Available: \url{https://qisdk.softbankrobotics.com/sdk/doc/pepper-sdk/ch4_api/conversation/reference/chat.html}
\bibitem{b32} Wikipedia contributers. (2022) Uncanny valley [Online]. Available: \url{https://en.wikipedia.org/wiki/Uncanny_valley}

\end{thebibliography}
%


\end{document}